\documentclass[suppldata]{interact}

\usepackage{epstopdf}
\usepackage{subfigure}

\usepackage{hyperref}
\theoremstyle{plain}

\theoremstyle{definition}

\theoremstyle{remark}

\usepackage[doi=true, isbn=false, url=false, eprint=false, maxnames=25]{biblatex}
\usepackage{graphicx}
\usepackage{booktabs}
\usepackage{caption} 
\usepackage{footmisc}

\usepackage{enumitem}

\usepackage{xspace}
\newcommand\ie{i.\,e.\xspace}
\newcommand\eg{e.\,g.\xspace}

\begin{document}


\title{pyWATTS: Python Workflow Automation Tool for Time Series}

\author{
\name{B. Heidrich\textsuperscript{a}, A. Bartschat\textsuperscript{a}, M. Turowski\textsuperscript{a}, O. Neumann\textsuperscript{a}, K. Phipps\textsuperscript{a}, S. Meisenbacher\textsuperscript{a}, K. Schmieder\textsuperscript{a}, N. Ludwig\textsuperscript{a}\textsuperscript{b}, R. Mikut\textsuperscript{a}, V. Hagenmeyer\textsuperscript{a}}
\affil{\textsuperscript{a}Institute for Automation and Applied Informatics (IAI), Karlsruhe Institute of Technology (KIT), Karlsruhe, Germany; \textsuperscript{b}Cluster of Excellence ``Machine Learning: New Perspectives for Science'', University of Tübingen, Germany}
}

\maketitle

\begin{abstract}
Time series data are fundamental for a variety of applications, ranging from financial markets to energy systems. Due to their importance, the number and complexity of tools and methods used for time series analysis is constantly increasing. However, due to unclear APIs and a lack of documentation, researchers struggle to integrate them into their research projects and replicate results. Additionally, in time series analysis there exist many repetitive tasks, which are often re-implemented for each project, unnecessarily costing time.
To solve these problems we present \texttt{pyWATTS}, an open-source Python-based package that is a non-sequential workflow automation tool for the analysis of time series data. \texttt{pyWATTS} includes modules with clearly defined interfaces to enable seamless integration of new or existing methods, subpipelining to easily reproduce repetitive tasks, load and save functionality to simply replicate results, and native support for key Python machine learning libraries such as \texttt{scikit-learn}, \texttt{PyTorch}, and \texttt{Keras}.
\end{abstract}

\begin{keywords}
Time Series Analysis; Python; Workflow Automation; Machine Learning; Pipeline
\end{keywords}

\section{Introduction}

In many areas, time series data are the most prominent form of data collected. In contrast to other sequential data such as speech data, time series data are not only ordered, but the time stamp associated with the observation might also have explicit information. For example, looking at energy time series, the demand at a specific time step depends on calendar-based information such as the day of the week or the season. Generally, time series analysis uses various algorithms from statistics to deep learning to answer questions about time-dependent systems.

Although more and more code from researchers focusing on time-dependent data is publicly available, there is still a need for respective tools. These tools should allow automating the workflow in time series analysis and an easy integration of new research approaches with third-party code. Automating the workflow is necessary, since many preprocessing tasks are repetitive, such as accounting for seasonality, adding calendar-based features, or detecting and imputing missing values. As a result of the lacking tools, researchers often re-implement these repetitive tasks at the unnecessary expense of time. Moreover, it is challenging to integrate new or alternative approaches into existing code workflows and, although the push towards open science increases the importance of reproducibility, it is often difficult to replicate earlier experimental results. Thus, any tool to aid researchers in automated time series analysis needs to focus on two features: re-usability and reproducibility of existing and new code. 

Several factors currently prevent good integration and re-usability of publicly available code for time series analysis. For example, most authors only publish their proposed new algorithm or method, excluding any steps necessary to prepare the data. Using their code then entails re-writing the required preprocessing method. Additionally, interfaces are hardly ever defined and basic unit-testing is often non existent, which regularly leads to re-implementation being the only quick and attainable solution. Regarding reproducibility, some of the issues include platform-dependent code, no information on parameter settings, or insufficient description on the order in which function or scripts need to be executed, making it almost impossible to reproduce results.

A remedy to the issues mentioned is workflow automation using pipelines and modules. In a pipeline, one can define the workflow, \ie the exact order in which several modules, each including a method, are run to achieve a specific result. No matter if we wish to use or reproduce code with pipelines, the steps needed to reach a specific result can be non-sequential. For example, we might wish to run parts of the code in parallel, have branching and merging pathways in the workflow, or even condition-dependent paths. Furthermore, the modules have a clearly defined structure which allows simple integration of new or alternative methods into an existing workflow.

Current Python tools which allow the realisation of pipelines are, for example, \texttt{scikit-pipeline}\footnote{\href{https://scikit-learn.org/stable/modules/generated/sklearn.pipeline.Pipeline.html}{https://scikit-learn.org/stable/modules/generated/sklearn.pipeline.Pipeline.html}} \cite{sklearn11} and \texttt{river}\footnote{\href{https://github.com/online-ml/river}{https://github.com/online-ml/river}} \cite{2020river}. While \texttt{scikit-pipeline} is part of the package \texttt{scikit-learn} \cite{sklearn11}, \texttt{river} is a merger of \texttt{CremeML} and \texttt{Sk-multiflow}. However, both tools only allow linear execution of modules, where neither parallel nor conditional execution is possible. Only the package \texttt{baikal}\footnote{\href{https://github.com/alegonz/baikal/}{https://github.com/alegonz/baikal/}} provides non-sequential pipelines inherited from \texttt{scikit-learn}. It is based on wrappers for \texttt{scikit-learn} modules, where each module has to be wrapped individually. Therefore, it is rather tedious. Furthermore, it does not allow integration of other libraries such as \texttt{PyTorch}  \cite{Paszke2019} and \texttt{Keras} \cite{chollet2015}, which are useful for deep learning-based time series approaches. Additionally, \texttt{baikal} aims to combine several \texttt{scikit-learn} modules such that they work as one module and thus focuses on the model creation only.


In the present paper, we introduce \texttt{pyWATTS}, an open-source Python-based package that provides a non-sequential workflow automation tool for the analysis of time series data. In contrast to \texttt{baikal}, \texttt{pyWATTS} includes generic wrappers for libraries such as \texttt{scikit-learn}, \texttt{PyTorch}, and \texttt{Keras}, allows the pipelines to have conditions, and is able to visualise intermediate results. Summarising the key features, \texttt{pyWATTS}
\begin{itemize}
    \item is a platform-independent solution to implement workflows from start to finish using pipelines. Thereby, time series experiments can be performed in an organised manner and in any environment that supports \texttt{pyWATTS}.
    \item enables re-usability through subpipelining. Any useful part of a time series experiment, \eg preprocessing, can be defined as a subpipeline and integrated into other pipelines without further adaption and independently of the original experiment. 
    \item allows saving and loading of any given pipeline configuration to reproduce results at a later date. 
    \item enables simple integration of new research approaches through a plug-and-play style environment where modules implemented in \texttt{pyWATTS} can be exchanged seamlessly between pipelines through a modular architecture with data handling through \texttt{xarray} \cite{Hoyer2017}.
    \item includes a clear API of the modules, \ie transform and fit methods, ensuring that pipelines within \texttt{pyWATTS} are adaptable and that modules can easily run on multiple data sets, at different points in a pipeline, and in various pipelines.
    \item  allows using different modules for the same part in the pipeline such that a condition mechanism decides which module is executed depending on the characteristics of the applied data.
    \item takes \texttt{pandas DataFrame}\cite{mckinney2010, reback2020} or \texttt{xarray Dataset} as input, allowing users to flexibly read the data from any source (file, database, website) with their method of choice.
    \item is able to use callbacks, \eg for visualising, analysing, and writing the intermediate results of modules. 
\end{itemize}

To the best of our knowledge, \texttt{pyWATTS} is the first tool to automate time series analysis workflows using non-sequential pipelines in this form. The remainder of the paper is structured as follows. We first introduce the implementation and architecture of \texttt{pyWATTS} before providing an overview on the availability and re-usability. We conclude with an outlook on ongoing and future projects that use \texttt{pyWATTS}.

\section{Implementation and architecture}

Implementing the features mentioned above requires a careful design of the architecture. In this section, we, therefore, describe \texttt{pyWATTS}' architecture and implementation. 

\begin{figure}
    \centering
    \includegraphics[width=.9 \textwidth]{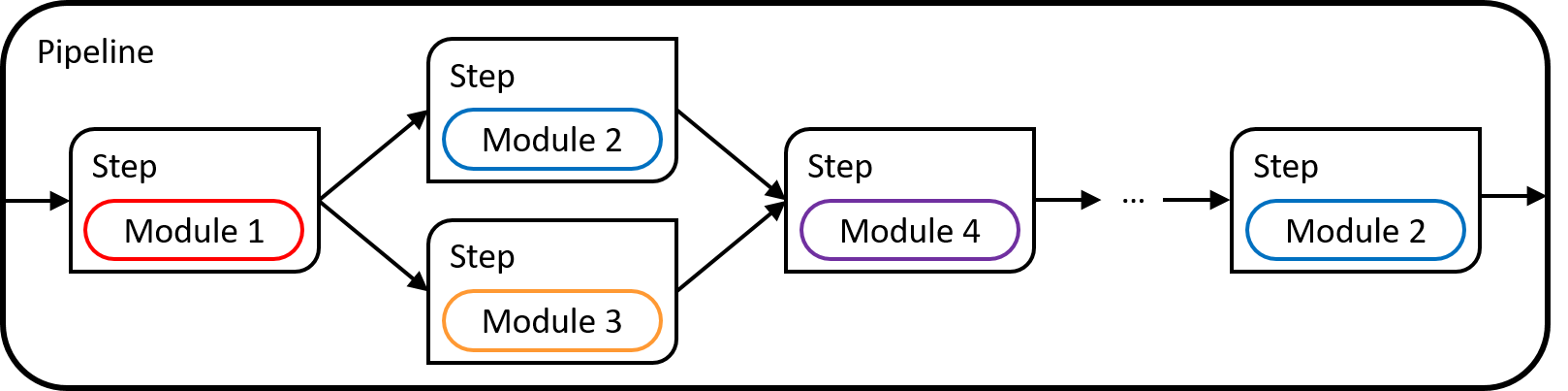}
    \caption{\texttt{pyWATTS} uses the three classes \emph{pipeline}, \emph{step}, and \emph{module} to realise non-sequential workflows.}
    \label{fig:pipeline}
\end{figure}

The \texttt{pyWATTS} package is written in the programming language Python\footnote{\url{https://www.python.org/}}. To realise non-sequential workflows, it uses three classes as illustrated in Figure~\ref{fig:pipeline}; a \emph{pipeline}, representing the workflow as a graph, a \emph{step} which represents a node in the graph referencing its dependencies with edges, and a \emph{module}, representing the algorithm running in a step. We introduce these three classes in more detail in the following.

Every algorithm used in \texttt{pyWATTS} is implemented with the module class. \texttt{pyWATTS} distinguishes between algorithms requiring training and algorithms that can be applied without training. For both types, the module class's \emph{transform} method must be implemented to apply the algorithm. For algorithms requiring training, we additionally have to implement the module class's \emph{fit} method that defines the training of the machine learning model. Modules must also include methods to save and load all information necessary for executing the module. 

In general, the implementation of modules follows the concepts introduced by \texttt{scikit-learn} \cite{sklearn11}. \texttt{pyWATTS} itself provides a comprehensive library of algorithms, as listed in \autoref{tab:pyWATTS_modules}. The currently available modules implement utile algorithms for time series analytics and serve as a guideline to implement further modules. \texttt{pyWATTS} also provides special modules, called wrappers, to seamlessly integrate existing algorithms and models from \texttt{scikit-learn} \cite{sklearn11} or deep learning models implemented in \texttt{Keras} \cite{chollet2015}, or \texttt{PyTorch} \cite{Paszke2019},

\begin{table}%
    \centering
    \footnotesize
    \caption[\texttt{pyWATTS} module library.]{The library of \texttt{pyWATTS} contains several utile algorithms when dealing with time series.}
\begin{tabular}{lp{25em}}
\toprule
\textbf{Module name} & \textbf{Description} \\
\midrule
Calendar extraction & Extracts or extends a time-series with calendar information such as weekdays or holidays \\
\addlinespace[0.5em]
Change Direction & Extracts if the change is positive or negative for each time point in a time series \\
\addlinespace[0.5em]
Clock Shift & Shifts the data with a certain offset \\
\addlinespace[0.3em]
Differentiate & Calculates the n-th order difference of a time series \\
\addlinespace[0.5em]
Linear Interpolator & Creates a linear interpolation \\
\addlinespace[0.5em]
Missing Value Detector & Detects missing values such as "NaN" \\
\addlinespace[0.5em]
Resampler & Reduces or increases the temporal resolution of a given time series \\
\addlinespace[0.5em]
Rolling Mean & Calculates a rolling mean over a specific window size \\
\addlinespace[0.5em]
RMSE Calculator & Calculates the Root Mean Squared Error (RMSE) \\
\addlinespace[0.5em]
Sampler & Creates samples with a specified sample size \\
\addlinespace[0.5em]
Trend Extraction & Extracts a trend specified by a period and a length \\
\addlinespace[0.5em]
Sklearn Wrapper & Wraps machine learning modules from the \texttt{scikit-learn} library \\
\addlinespace[0.5em]
Keras Wrapper & Wraps deep learning neural networks implemented in the \texttt{Keras} library \\
\addlinespace[0.5em]
PyTorch Wrapper & Wraps deep learning neural networks implemented in the \texttt{PyTorch} library \\
\bottomrule
\end{tabular}%

    \label{tab:pyWATTS_modules}%
\end{table}

Given a module, \texttt{pyWATTS} creates one or multiple steps\footnote{A step contains zero or one module. A module can be used in multiple steps.}. The step class organises the execution of the pipeline. A step collects and merges the results of its dependencies, calls the fit and transform method of its module, and provides its output to the pipeline for the subsequent steps. Moreover, a step can execute callbacks defined by the user, \eg for visualising, analysing, and writing the module's intermediate result. Furthermore, a step controls the execution of the module based on conditions defined by the user.

The pipeline class organises the steps in nodes and creates a graph, where every step input is represented as an edge. This graph supports branching and merging of paths and is used to define the execution order of the steps. This way, all previous steps represented as dependencies have to be successfully executed before the current step itself is executed. The pipeline also serves as the interface to the user and provides control commands. These commands include training and executing a pipeline, as well as saving and loading the whole pipeline.

\begin{figure}
    \centering
    \includegraphics[width=.5 \textwidth]{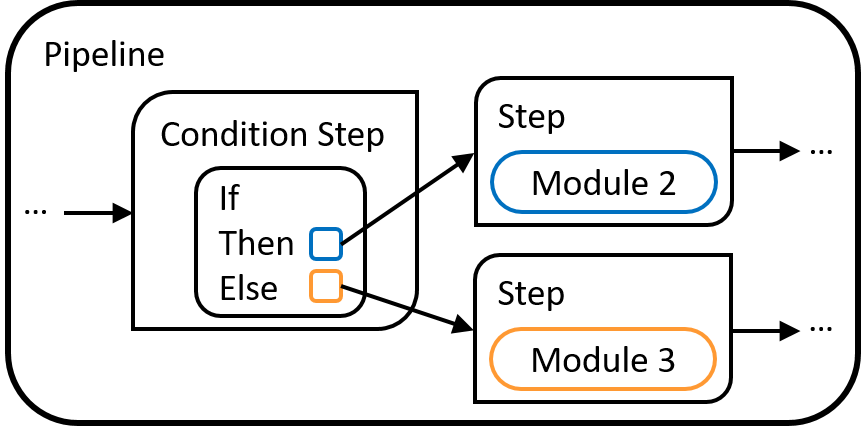}
    \caption{\texttt{pyWATTS} uses condition steps with "if-then-else" for conditional branching in pipelines.}
    \label{fig:conditionals}
\end{figure}

\begin{figure}
    \centering
    \includegraphics[width=.9 \textwidth]{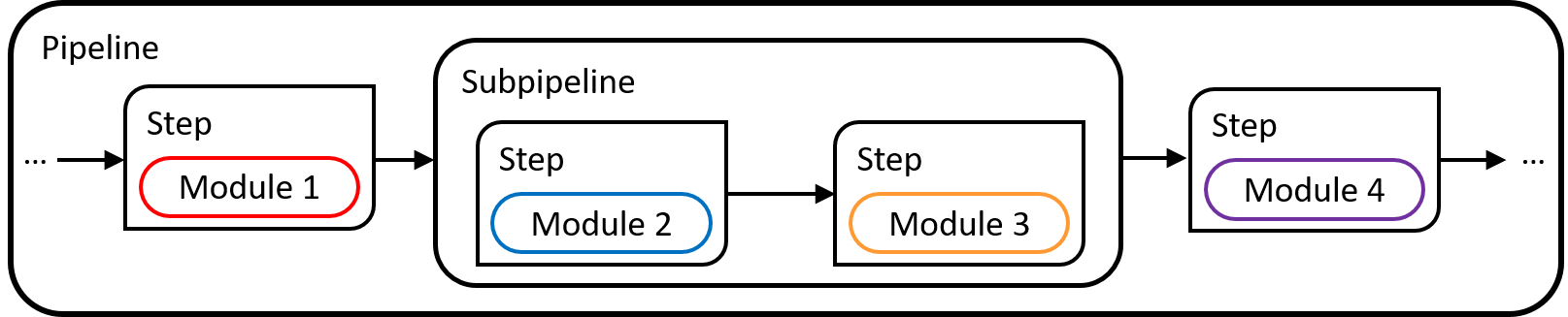}
    \caption{\texttt{pyWATTS} makes use of subpipelines to easily structure and name parts of the workflow.}
    \label{fig:subpipelines}
\end{figure}

Based on the mentioned three classes, \texttt{pyWATTS} implements the following three functionalities for an easy structuring and flexible application of pipelines:
\begin{description}
    \item[Batch/online learning:] By specifying that the pipeline processes only one time step at a time, the pipeline can be executed iteratively.
    \item[Conditional branching:] Depending on the applied data, condition steps with "if-then-else" can be used to select different paths of the pipeline for execution (see Figure~\ref{fig:conditionals}).
    \item[Subpipelines:] Grouping steps of the pipeline in subpipelines allows an easy structuring and naming of certain parts of the workflow (see Figure~\ref{fig:subpipelines}).
\end{description}

\section{Quality control}

For quality control, we apply comprehensive testing to \texttt{pyWATTS}, use programming guidelines and code reviews, as well as provide up-to-date documentation and examples. For automated testing, we use GitHub Actions. All core classes implementing the main functionality of the pipeline and the steps are tested with unit tests. Furthermore, unit tests cover the wrappers and the modules of the library. To ensure the correct interaction of steps and modules as well as saving and loading of whole workflows, exemplary pipelines are implemented as integration tests.

Furthermore, we follow programming guidelines to ensure high-quality source code. These guidelines define the naming of branches and code conventions as well as prescribe the use of linting software such as pylint\footnote{\url{https://www.pylint.org/}}. Additionally, the guidelines require developers to implement tests for each module and to use loggers with appropriate logging messages. Finally, the guidelines demand developers to use type annotations for arguments, variables, and return values. In the GitHub Actions, we automatically check whether code conventions are met using flake8\footnote{\url{https://gitlab.com/pycqa/flake8}}. To ensure compliance with all guidelines, we additionally perform manual code reviews on pull requests. Maintainers review pull requests concerning their correct operation, coverage through tests, and compliance to programming guidelines and code conventions before the pull requests are merged into the master branch. 

Lastly, we maintain an up-to-date documentation. Based on the annotated source code and restructured text files, the documentation\footnote{The \texttt{pyWATTS} Documentation is available at \url{https://pywatts.readthedocs.io/en/latest/}.} of \texttt{pyWATTS} is automatically generated using sphinx\footnote{\url{https://www.sphinx-doc.org/en/master/}} and readthedoc\footnote{\url{https://readthedocs.org/}.}. 

Besides serving as integration tests, the provided examples introduce new users to \texttt{pyWATTS} and its features and support them in creating working pipelines in \texttt{pyWATTS}. In the following, we briefly describe the provided examples, which are detailed in the documentation.

\begin{itemize}
    \item To prevent fundamental errors during the creation of a pipeline, a simple example explains how one can create a pipeline for electrical load forecasting and how one can add modules such as the Calendar Extraction to the pipeline.
    \item To test the functionality of the condition mechanism, we provide an example that changes the method for electrical load forecasting depending on day-time and night-time.
    \item Advanced examples aim to avoid mistakes in the application of deep learning frameworks in \texttt{pyWATTS}. In the examples using \texttt{Keras} \cite{chollet2015} or \texttt{PyTorch} \cite{Paszke2019}, the pipelines train simple deep learning models.
\end{itemize}

\section{Availability}

\subsection*{Operating system}
Platform independent

\subsection*{Programming language}
Python

\subsection*{Additional system requirements}
\texttt{pyWATTS} is designed to perform various time series analysis tasks on data sets of arbitrary size. Therefore, hardware requirements depend on the size of the data set and the task being performed.

\subsection*{Dependencies}
The core \texttt{pyWATTS} dependencies are the following:
\begin{itemize}[noitemsep]
    \item scikit-learn -- 0.23.2
    \item cloudpickle -- 1.6.0  
    \item xarray -- 0.16.1 
    \item numpy -- 1.19.2 
    \item pandas -- 1.1.5 
    \item matplotlib -- 3.3.2 
    \item tensorflow -- 2.3.1 
    \item workalendar -- 12.0.0
    \end{itemize}
Dependencies required for development purposes comprise the following:
\begin{itemize}[noitemsep]
    \item pytest -- 6.1.1 
    \item sphinx -- 3.2.1 
    \item pylint -- 2.6.0
    \item pytest-cov -- 2.10.1
\end{itemize}

\subsection*{Software location:}

{\bf Archive}

\begin{description}[noitemsep,topsep=0pt]
	\item[Name:] Zenodo
	\item[Persistent identifier:] \href{https://doi.org/10.5281/zenodo.4637197}{https://doi.org/10.5281/zenodo.4637197}
	\item[Licence:] MIT Licence\footnote{\href{https://opensource.org/licenses/MIT}{https://opensource.org/licenses/MIT}\label{MIT-license}}
	\item[Publisher:]  Zenodo
	\item[Version published:] 0.1.0
	\item[Date published:] 25.03.2021
\end{description}

{\bf Code repository} GitHub

\begin{description}[noitemsep,topsep=0pt]
	\item[Name:] \texttt{pyWATTS}
	\item[Persistent identifier:] \href{https://github.com/KIT-IAI/pyWATTS}{https://github.com/KIT-IAI/pyWATTS}
	\item[Licence:] MIT Licence 
	\item[Date published:] 25/09/2020
\end{description}

\subsection*{Language}
English

\section{Reuse potential}


Due to the architecture and the modular structure of \texttt{pyWATTS}, anyone who wishes to analyse time series can use \texttt{pyWATTS} out-of-the-box. It enables the users to easily select the modules and determine the pipeline structure relevant for their specific use case, such as forecasting.
Additionally, the possibility to save and load pipelines together with the platform-independence of \texttt{pyWATTS}, allows easy reproduction of research results.
Moreover, common Python-based machine learning libraries can be used within \texttt{pyWATTS}. For example, we provide wrapper modules for \texttt{scikit-learn} \cite{sklearn11}, \texttt{Keras} \cite{chollet2015}, and \texttt{PyTorch} \cite{Paszke2019} to allow the inclusion of the available functions.

Moreover, \texttt{pyWATTS}' users are supported by comprehensive documentation for its core structure and the individual modules as well as detailed examples. In case of questions, the core developer team can also be contacted with the help of GitHub issues or the \texttt{pyWATTS} contact email address. The generous MIT license\footref{MIT-license} allows research, commercial and non-commercial use, and development of the package as either an anonymous user, private developer or publicly contributing developer. All users can stick to the existing modules and pipelines, extend them based on known or unknown issues, or create new modules and pipelines. Whether any changes to the modules are made locally or through the public repository is up to the user to decide.

The developer team, for example, wants to use \texttt{pyWATTS} in various research applications in the future. For preprocessing, we plan to extend \texttt{pyWATTS} with the Copy Paste Imputation of missing values for energy time series as described in \cite{Weber2021}. We also plan to use \texttt{pyWATTS} for time series forecasting, \eg by using Profile Neural Networks \cite{Heidrich.2020}. Furthermore, we intend to extend \texttt{pyWATTS} for the insertion of typical anomalies in energy time series to have data sets with ground truth for anomaly detection and anomaly handling. To generate realistic synthetic energy time series, we also aim to use \texttt{pyWATTS}. An interface for pipeline tuning and selection will further assist in automating the iterative design process. Lastly, we want to deploy \texttt{pyWATTS} as an execution environment in the research infrastructure Energy Lab 2.0 \cite{Hagenmeyer.2016}.

Putting it all in a nutshell, \texttt{pyWATTS} provides an extendable framework for automating time series analysis workflows. It uses comprehensible pipelines and is able to integrate established statistical, machine learning, and deep learning frameworks. Thus, \texttt{pyWATTS} makes it easy to develop, adapt, and reproduce pipeline-based experiments for energy time series analysis.

\section*{Acknowledgements}


We thank Simon Waczowicz for the valuable input on the concept of \texttt{pyWATTS}.

\section*{Funding statement}


This project is funded by the Helmholtz Association's Initiative and Networking Fund through Helmholtz AI, the Helmholtz Association under the Program ``Energy System Design'', the Joint Initiative ``Energy System Design - A Contribution of the Research Field Energy'', the Helmholtz Metadata Collaboration, and the German Research Foundation (DFG) as part of the Research Training Group 2153 ``Energy Status Data: Informatics Methods for its Collection, Analysis and Exploitation'' and under Germany’s Excellence Strategy – EXC number 2064/1 – Project number 390727645.
\newpage
\printbibliography

\end{document}